\documentclass[conference]{IEEEtran}

\usepackage{cite}
\usepackage{amsmath,amssymb,amsfonts}
\usepackage{algorithmic}
\usepackage{graphicx}
\usepackage{textcomp}
\usepackage{xcolor}
\def\BibTeX{{\rm B\kern-.05em{\sc i\kern-.025em b}\kern-.08em
    T\kern-.1667em\lower.7ex\hbox{E}\kern-.125emX}}
\begin{document}

\title{Multi-Objective Risk Assessment Framework for Exploration Planning Using Terrain and Traversability Analysis\\
}

\author{Riana Gagnon Souleiman, Vivek Shankar Varadharajan, Giovanni Beltrame
\thanks{The authors are with the MIST Lab, Polytechnique Montreal, Canada. Contact: {\tt riana.gagnon-souleiman@polymtl.ca}}}

\maketitle

\begin{abstract}
Exploration of unknown, unstructured environments—such as in search and rescue, cave exploration, and planetary missions—presents significant challenges due to their unpredictable nature. This unpredictability can lead to inefficient path planning and potential mission failures. We propose a multi-objective risk assessment method for exploration planning in such unconstrained environments. Our approach dynamically adjusts the weight of various risk factors to prevent the robot from undertaking lethal actions too early in the mission. By gradually increasing the allowable risk as the mission progresses, our method enables more efficient exploration. We evaluate risk based on environmental terrain properties, including elevation, slope, roughness, and traversability, and account for factors like battery life, mission duration, and travel distance. Our method is validated through experiments in various subterranean simulated cave environments. The results demonstrate that our approach ensures consistent exploration without incurring lethal actions, while introducing minimal computational overhead to the planning process. 
\end{abstract}

\section{Introduction}
The use of unmanned ground vehicles (UGVs) for exploration in large unstructured environments has become increasingly popular but poses many challenges due to the uncertainty of the terrain such as surface friction, slopes, obstacles and much more. The European Space Agency (ESA) currently has plans to explore caves and underground lava tubes on the moon using autonomous rovers \cite{esalunar}. Prior knowledge of the environment is required to efficiently navigate these environments to mitigate the total failure of the system due to tipping or getting stuck \cite{Fankhauser2018ProbabilisticTerrainMapping}. Many other considerations must be made during exploration, such as mission time, distances, power consumption and exploration gain as these are all crucial components of exploration missions. 
\par 
Heterogeneous ground robots are equipped with diverse capabilities, leading to varying responses to different physical terrain characteristics. Wheeled robots are typically designed for efficient energy utilization but struggle with maneuverability on soft and deformable soils. In contrast, tracked mobile robots exhibit high maneuverability on rough terrain but require more power due to their heavier weight \cite{RobotSurvey}. These differences should be taken into account during planning to adequately assess the risk of traversing highly unstructured terrain, as some robots may struggle with certain tasks while others are highly capable.
\par
This paper introduces a multi-objective risk assessment pipeline for exploration planning that integrates terrain analysis into the decision-making processes. The proposed method evaluates terrain characteristics, such as slope and roughness, and balances these risks with other mission-critical objectives like battery life, mission duration, and path distance. By dynamically adjusting these factors, the planner improves the overall robustness of robotic exploration in unstructured environments, ensuring both safety and efficiency throughout the mission. The paper will begin with a discussion on related works in exploration, path planning and terrain analysis. The methodology section will discuss the methods used for multi-objective optimization and risk assessment. Finally, results and conclusions will discuss the final results of the integration of the risk assessment framework into an exploration planner.

\begin{figure}
    \centering
    \includegraphics[width=1\linewidth]{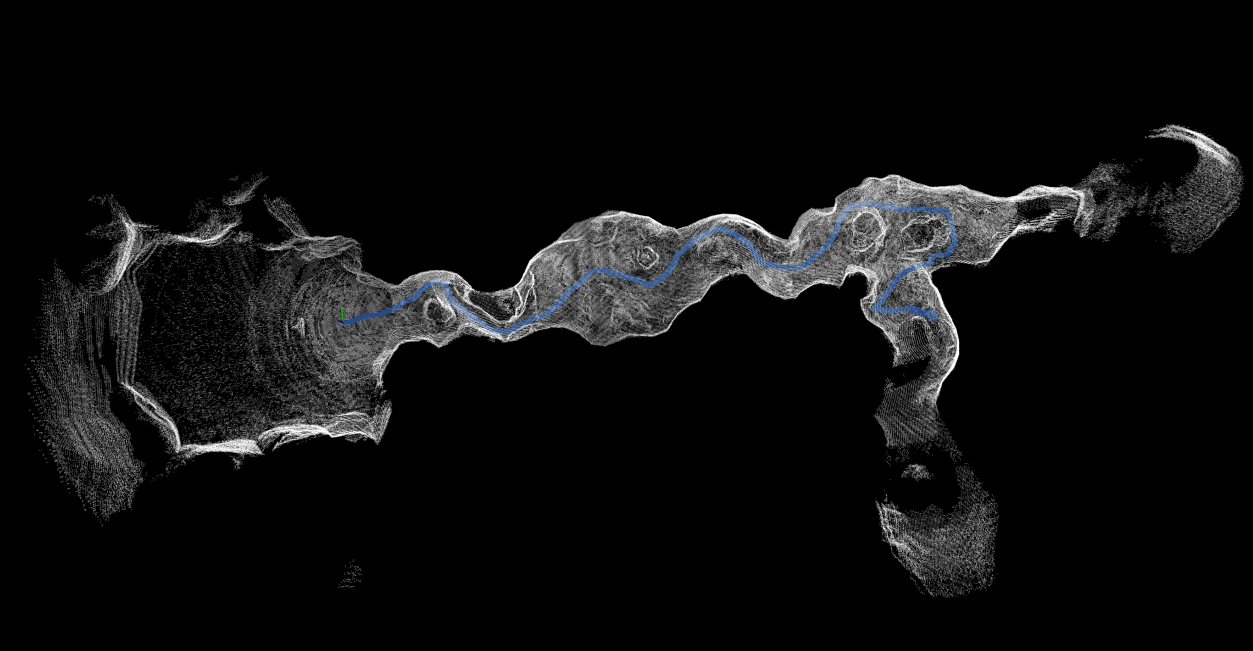}
    \caption{Trajectory of the agent in DARPA Subterranean Cave World 1 Environment utilizing the risk assessment framework while exploration planning using terrain analysis from onboard sensor measurements}
    \label{fig:trajectory}
\end{figure}
\section{Related Work}

Exploration planning is an open problem in robotics that aims to answer how to most effectively explore an area while considering different challenges such as robot malfunction and uncertainty, environmental factors, and path optimality \cite{compCPP, morrell2024addendum, hudson2021heterogeneous}. Exploration planning couples the problem of online path planning with finding the best set of actions for agents to take to most effectively explore an unknown environment. Exploration planners use volumetric gains as an objective to select the next goal state for the robot. Some works use techniques such as receding horizon \cite{recedeaerial} \cite{horizonNBV} to optimize decision-making over a continuous time horizon. Yamauchi~\cite{yamauchi1998frontier} proposed a method of exploration using frontiers, identifying regions on the boundary of unexplored and open space \cite{frontiers}.  Dang et al. proposed a graph based exploration planning method for subterranean environments utilizing local gain and global frontiers to optimize exploration gain \cite{gbplanner}. Works that use classic path planning algorithms like A* and Dijkstra's focus on planning optimization by choosing the shortest path to a goal  \cite{improveda} \cite{onlineDij}. There have been numerous studies on risk aware planning to mitigate risk exposure in different environments \cite{riskDORA} \cite{CvarQuad}. While many of these planning algorithms focus on performance of the agents in flat, structured environments, we wish to consider the terrain when planning for more robust performance in unstructured environments like caves and extraterrestrial planetary surfaces. The importance of traversability analysis and risk have been highlighted in the methods proposed within the Darpa Subterranean challenge~\cite{tranzatto2022cerberus, agha2021nebula, hudson2021heterogeneous}. Studies have considered energy costs through power consumption models. Sakayori and Ishigami proposed an energy-aware trajectory planner that considers wheel dynamics and terra mechanics to employ a cost function model for trajectory planning \cite{energyPlanner}. Daniel et al. proposed the Theta* algorithm, which uses classical optimization to save the energy consumption of the mobile robot by smoothing the paths generated by the A* algorithm \cite{Daniel_2010}. Some works have been done on risk-aware planning using terrain considerations \cite{riskprobs} \cite{slipslope} \cite{callike}. Fan et al. proposed a risk-aware Conditional Value at Risk (CVaR) planning framework that combines different sources of risk such as slip, tip, and collision. Many of these works focus on optimizing risks related to time, energy consumption, distance, and unstructured terrain separately. Although, we wish to combine these metrics and perform multi-objective optimization for exploration planning to ensure both safety and efficiency of the planner. We propose a risk assessment framework that considers risks from terrain, power consumption requirements, path distance, and mission time that will improve the robustness of exploration planner's to failure in harsh, unstructured environments.

\section{Methodology}
\subsection{Problem Formulation}
Let the exploration space be defined as $\mathcal{Q} \subset \mathbb{R}^3$, with $\mathcal{Q}_{trav} \subset \mathcal{Q}$ representing the traversable subspace, where agents can safely navigate and $\mathcal{Q}_{untrav} = \mathcal{Q}/\mathcal{Q}_{trav}$. An agent is described by its pose at time step $k$, denoted as $x_{k} = [p_{k}, r_{k}]$, where $p_{k} \in \mathbb{R}^3$ is the position and $r_{k} \in SO(3)$ is the orientation. 

The environment is modeled using a discretized voxel map, where each voxel $v \in \mathbf{M}$ represents a portion of the space $\mathcal{Q}$, and $\mathbf{M} \subseteq \mathcal{Q}$ is the set of all voxels with a resolution $m_{res}$. The subset $\mathbf{M}_{G} \subset \mathbf{M}$ represents the ground plane, which the robot observes through its line-of-sight sensor.

For each observed ground plane $\mathbf{M}_{G}$, we estimate the ground height as $h_i \in \mathbb{R}$, with a confidence interval $[h_{i,min}, h_{i,max}]$. This height data is used to construct the elevation map $\mathbf{G} \in \mathbb{R}^2$, a grid map containing height estimates at each cell $h_i \in \mathbb{R}$, $h_{i,min}, h_{i,max} \in \mathbb{R}$. Additional grid map layers, such as surface normal, roughness, and slope, are computed based on the elevation map. These layers are combined into a multi-layer grid map $\mathbf{G}^{L}$, where $\mathbf{G}^{l}$ represents the $l_{th}$ layer, and $g^{l}_{i}$ is the $i_{th}$ cell of the $l_{th}$ layer: $\mathbf{G}^{L} = \{ \mathbf{G}^{1}, \mathbf{G}^{2}, \dots, \mathbf{G}^{L} \}, \quad g^{l}_{i} \in \mathbf{G}^{l}$.

The path of the robot is denoted by:
$P_{0:N} = [x_{0}, x_{1}, x_{2}, \dots, x_{N}],$
where $N$ is the number of waypoints, and each $x_k$ represents the pose of the robot at time step $k$:
$x_{k} = [p_{k}, r_{k}], \quad p_{k} \in \mathbb{R}^3, \quad r_{k} \in SO(3).$


\textit{Problem Statement:} 
Given the initial pose $\mathbf{x}_0 $ of the robot and the multi-layer grid map $\mathbf{G}^{L}$ constructed from sensor observations, the objective is to compute a collision-free path $P$ that
maximizes the exploration and information gain of the voxel map $\mathbf{M}$ while minimizing risk, energy, traveled distance, and mission time.

The objective of maximizing the information gain can be defined as $\max(I(\mathbf{M}_{explored}))$
where $I(\mathbf{M}_{explored})$ is the information gain from the explored voxel map by the robot. 

Minimizing risk factors include minimizing the risk induced by terrain harshness computed through elevation map and roughness layers: 
\[
\min \sum_{k=1}^{N} R(x_k),
\]
where $R(x_k)$ is the risk cost at pose $x_k$ based on the terrain map.

The overall energy consumption of the robot is defined as $\min( E(P) )$ where $E(P)$ is the energy cost of the robot's path. The traveled distance and the time is minimized as follows $ \min(\text{Time}(P) + \text{Distance}(P) )$.

\subsection{Terrain Maps}

\begin{figure}[tbp]
\includegraphics[width=\columnwidth]{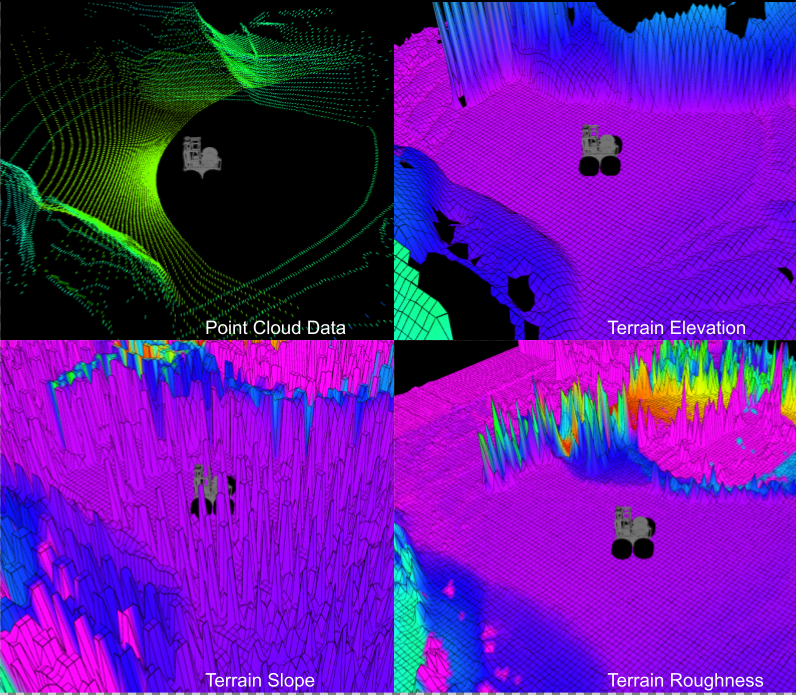}
\caption{Illustration of the sensor data and the various layers of the terrain map used to assign a risk factor to the robots at a certain position in the environment.}
 \label{fig:terrainmap}
\end{figure}

Terrain mapping involves processing onboard sensor measurements such as LiDAR pointcloud data and fusing it with proprioceptive state estimation to generate maps with physical terrain characteristic data such as elevation, slope, roughness, and surface normals. We use the bivariate cumulative distribution function (CDF) method proposed in~\cite{Fankhauser2014RobotCentricElevationMapping} to generate the elevation estimates from the sensor measurements in the robot's reference frame. Figure \ref{fig:terrainmap} shows an example of the various terrain maps used to compute the risk of the individual robots.

\subsection{Risk Assessment}

This work proposes a risk assessment pipeline to enhance multi-robot exploration planning in unknown environments. By leveraging environmental characteristics observed through onboard sensing, the pipeline generates collision-free trajectories that minimize traversal risks. Robot failures, such as those caused by difficult terrain or mechanical limitations, are mitigated by considering factors like terrain roughness, slope, and each robot's capabilities. The proposed method dynamically balances multiple objectives, risk, time, distance, and energy using real-time sensory feedback. By assigning dynamic weights to these objectives, the system adapts to mission progression, allowing it to prioritize objectives based on current conditions. This adaptive approach enables the mission to remain flexible, shifting focus between exploration speed, safety, and energy efficiency as the environment evolves.

\subsubsection{Collision Risk} 
Collision risk can be defined as the probability that an agent will collide with or traverse an area it is incapable of crossing without failure. The risk cost, $ R_{c} $, is determined based on the distance from the nearest obstacle. An inflation process is applied to propagate risk costs from untraversable terrain cells, where the cost decreases as the distance from these cells increases. The collision risk cost is captured by the following equation:
\begin{equation}
R_{c} = e^{(X_{dim} - d_{min})}
\end{equation}
where $X_{dim}$ is the maximum dimension of the agent, and $d_{min}$ is the shortest distance from the position of untraversable cell $x_{u} \in \mathcal{Q}_{untrav}$. The minimum distance $d_{min}$ can be defined as:
\[d_{min} = argmin(||x_k - x_{u}||)\]
where $ x_k $ is the agent's current position, and $ x_u $ represents the position of the nearest untraversable terrain cell.
The position of untraversable terrain, $ x_u $, is derived from the traversability terrain map. Specifically, it is the global position of the $i_{th}$ index in the grid map where the traversability value $ g^{trav}_i = 0 $ (indicating untraversable cells). Thus, as $ d_{min} $ decreases, $ R_{c} $ increases exponentially, and the collision risk is bounded within the range $[0, 1]$.

\subsubsection{Traversability}

A traversability cost $R_{t}$ can be quantified by the normalized sum of the slope and roughness of the terrain at a given pose $x_k$. The roughness is calculated as the absolute difference between the terrain's elevation map and a smoothed version of the same map. Roughness values are bounded between $[0, 0.1]$, with $0$ indicating smooth terrain and $0.1$ representing highly rough terrain. The slope is computed as the inverse cosine of the surface normal vector, which describes the steepness of the terrain.

To compute the overall traversability cost $R_{t}$, we use the following equation:
\[
R_{t} = \frac{\theta_{x_k}}{\theta_{max}} + \frac{r_{x_k}}{r_{max}},
\]
where $\theta_{x_k}$ is the slope at the robot's current pose $x_k$, $r_{x_k}$ is the roughness of the terrain at $x_k$, $r_{max} = 0.1$ is the maximum roughness value, and $\theta_{max}$ is the maximum slope angle the robot is capable of traversing, which is dependent on the robot's model and is typically specified by its manufacturer.

The above equation normalizes both terrain slope and roughness relative to their maximum allowable values, ensuring that the traversability cost reflects both aspects of the terrain's difficulty.

\subsubsection{Slip Risk}

Slip risk is largely dependent on the slope, friction, weight, and wheel surface area in contact with the ground. While tracked robots tend to be heavier and less energy efficient, their ability to maneuver soft and deformable terrain is largely due to their contact surface area \cite{RobotSurvey}. Less surface area in contact with the ground results in more pressure applied to a single point. To quantify the cost of slip risk, we calculate the total pressure an agent exerts on the surface where the force can be defined as the normal force such that:
\[F = m_{robot}*g(cos(\theta_{x_k}))\]
where $m_{robot}$ is the mass of the robot and $g$ is gravity. Now, the equation for the surface area of the wheels in contact with the ground is:
\[A = t_{wheel}*l_{contact}*N_{wheels}\]
Where $t_{wheel}$ is the thickness of the wheel, $l_{contact}$ is the length of the wheel contacting the ground, and $N_{wheel}$ is the number of wheels the agent has, i.e., 4 for rovers and 2 for tracked robots.
Finally, we have the equation for pressure:
\[Pr = \frac{m_{robot}*g(cos(\theta_{x_k}))}{t_{wheel}*l_{contact}*N_{wheels}}\]

Lastly, studies have shown that the slip ratio, which defines the wheel behavior by comparing actual forward velocity,$v_x$, to commanded velocity,$v_{ref}$, will increase as slope increases \cite{slipslope}. A high slope ratio value indicates more slip, with a value of 1 representing a complete slip state. Additionally, other studies have shown that the probability of slipping will increase as friction, $\mu$, decreases \cite{frictionslip}. To account for these factors, we will multiply our pressure value by a slope-to-friction ratio to accurately reflect a slipping value, such that:
\[s= \frac{Pr*tan(\theta_{x_k})}{\mu}\]

Modeling this relationship shows that as pressure and slope increase and friction decreases, $s$ increases with a maximum value at $\theta = 90^\circ$ with $\theta$ being bounded between [0,$180^\circ$]. To normalize $s$, take the maximum value for slope, $\theta_{max}$, as the max inclination traversable by the robot. When planning with multiple agents, $s_{max}$ will be the highest value among all agents. To define a cost value:

\[Pr_{max} = \frac{m_{robot}*g(cos(\theta_{max}))}{t_{wheel}*l_{contact}*N_{wheels}}\]
\[s_{max} =\frac{Pr_{max}*tan(\theta_{max})}{\mu}\ \]

Finally, slipping risk can be defined as:
\[R_{s} = \frac{s}{s_{max}}\]

\subsubsection{Final Risk Cost}

Once our values for traversability, slope, and collision are calculated, we sum the values together to get a risk value at time step $k$:
\[R_k = R_{c}+R_{t}+R_{s}\]

As to not penalize longer paths with more time steps, $N$, we average the accepted risk over a path taken, $P_{0:N}$, so we have:
\[R_{total} = \frac{\Sigma^{N}_{k=0} R_k}{N}\]

\subsection{Energy}

Battery considerations are important for successful completion of missions. Considering power requirements for a given path is important so as to not overexert the system in order to ensure the environment can be adequately explored. Choosing energy efficient paths such as going around a hill rather than consuming the extra power needed to climb, may allow us to cover more area over a shorter duration by prolonging mission times before needing to return home to charge. To assess the power requirements for a given path, we will use the simple relationship:
\[Power = Force \cdot Velocity\]

The force can be defined as the sum of the forces acting against the forward motion of the wheeled robot, which are rolling resistance $F_r$ and gradient resistance $F_p$, with:
\begin{align*}
F_r &= mgcos(\theta_k)       &  F_p &= mgsin(\theta_k)  
\end{align*}

so utilizing the set max forward velocity of the agent:
\[Power = v_{max}(F_r+F_p)\]

Negative inclination checks should be done to account for the drop in power consumption needed when descending a hill, thus if a robot is traveling down a slope:
\[Power = v_{max}(F_r - F_p)\]
This value will increase as slope and mass of the robot increases factoring in the energy consideration for different robots. 

\subsection{Distance and Time}

Distance and time will be defined by the path where distance is the length of the path and time is time taken to traverse a path, $P_{0:N}$ where:
\begin{equation}
Distance = \Sigma^{N-1}_{k=0} ||x_k - x_{k+1}||^2
\end{equation}
and 
\[time = Distance/v_{max}\]

These are important factors to consider when restricted by mission time as well as specific max travel distances for different robots.

\subsection{Gain}

Exploration gain is defined as the expected cumulative unmapped volume that a sensor could perceive \cite{gbplanner} and can be generally represented by the expected information gain by integrating over all possible observations $\hat{z}$ such that:
\[E[I(a_t)] = \int_z p(\hat{z} | a_t, x_t)I(\hat{z},a_t)d\hat{z} \]

Where $a_t$ is the action taken and $x_t$ is the state of the robot at time t, as detailed in~\cite{infotheory}. The expected information gain is crucial to efficient exploration as it will dictate the set of actions the agent should take to best explore an unknown environment. While we aim to minimize the risk, energy, time, and distance metrics, the overarching goal is to decide whether the risk is worth the reward through multi-objective optimization.

\subsection{Multi-Objective Optimization}

Multi-objective decision making relies on weighting the importance of each asset. The chosen method for decision making is the VIKOR method \cite{VIKOR}, which will reflect a compromised solution by providing a maximum group utility, $S$, and individual regret, $R$, for each alternative. The alternatives, $i$, are defined as every decision option, i.e every feasible path to take, while criterion ,$j$, represents each objective(risk, time, distance, energy).
The group utility, $S$, and regret, $R$, at alternative, $i$, is defined as:

\begin{align*}
S_i &= \sum_{j=1}^n w_j\frac{f^*_j-f_{ij}}{f^*_j-f^-_j}  &  R_i &= max(w_j\frac{f^*_j-f_{ij}}{f^*_j-f^-_j}) 
\end{align*}

where $f^-_j = max(f_{ij})$ and $f^*_j = min(f_{ij})$ for $i = 1:m$ alternatives. The variable $w_j$ represents the chosen weight given to each criterion. Since we have 4 criterion we will have 4 different weights, $[w_{risk}, w_{distance}, w_{time}, w_{energy}]$. We want to dynamically set these weightings based on current mission information and constraints. For time it will depend on the ratio of elapsed time to mission time, energy will be scaled by the remaining battery life, and distance is the ratio of distance traveled to the specified max distance of the agent. So finally the time weight will be:
\[w_{time} = \frac{t_{elapsed}}{t_{mission}}\] 

The distance weight will be:
\[w_{distance} = \frac{d_{traversed}}{d_{max}}\]
Furthermore the energy weight is defined as:
\[w_{energy} = \frac{1}{t_{battery}}\]
We assume that we want to give initial priority as well as constant consideration to risk, so risk weight can be defined as:

\[w_{risk} = 1 - \frac{w_{distance}+ w_{time}+ w_{energy}}{3}\]

To calculate the final VIKOR value, $Q$, we use the equation:
\[Q_i = \upsilon\frac{S_i - S_{star}}{S^--S^*} + (1-\upsilon)\frac{R_i - R_{star}}{R^--R^*}\]

where $S^* = min(S_i)$, $S^- = max(S_i)$, $R^* = min(R_i)$, $R^- = max(R_i)$, and $\upsilon$ is the weight of majority. As the overall objective is maximize the information gain, $G$, we will take:
\[Q_{gain} = \frac{G_i - G_{min}}{G_{max} - G_{min}}\]
Lastly, to weigh the risk to reward, we wish to take the path P such that:
\[P = max(Q_{gain} - Q_{gain}*Q)\]

By integrating this pipeline into an exploration planner, multi-objective risk assessment can be completed to improve the robustness of the planner.

\section{Simulations}
\subsection{Experimental Setup}

The proposed risk-aware planning approach was integrated into the gbplanner2 Exploration Planner \cite{gbplanner}. The integration utilized the ground robot model and specifications provided by gbplanner2. To evaluate the effectiveness of the risk assessment during planning, we compared the enhanced risk-aware planner against the baseline gbplanner2. Simulations were conducted using two different environmental models of the Darpa Subterranean world. Specifically, ten runs of missions, each lasting 600 seconds, were performed in both cave models, with and without the risk-aware planning pipeline.
For the larger Darpa Cave World 1 model, additional runs of 1800 seconds were conducted, with ten repetitions of the experiment performed in this extended scenario.


\subsection{Results}

The experimental results highlight the impact of weighting risk versus reward and underscore the importance of incorporating risk considerations into planning for navigation in unstructured, unknown environments. When attempting to constrain risks to an allowable range for the agent’s traversal, the available path options become significantly constrained. This is due to the inherent nature of most paths in such environments, which often require a certain degree of risk to navigate successfully. Thus, while managing risk is crucial, it can also limit the feasible paths and potentially restrict exploration opportunities. 
\par
\begin{figure}[h!]
\includegraphics[width=8.6cm]{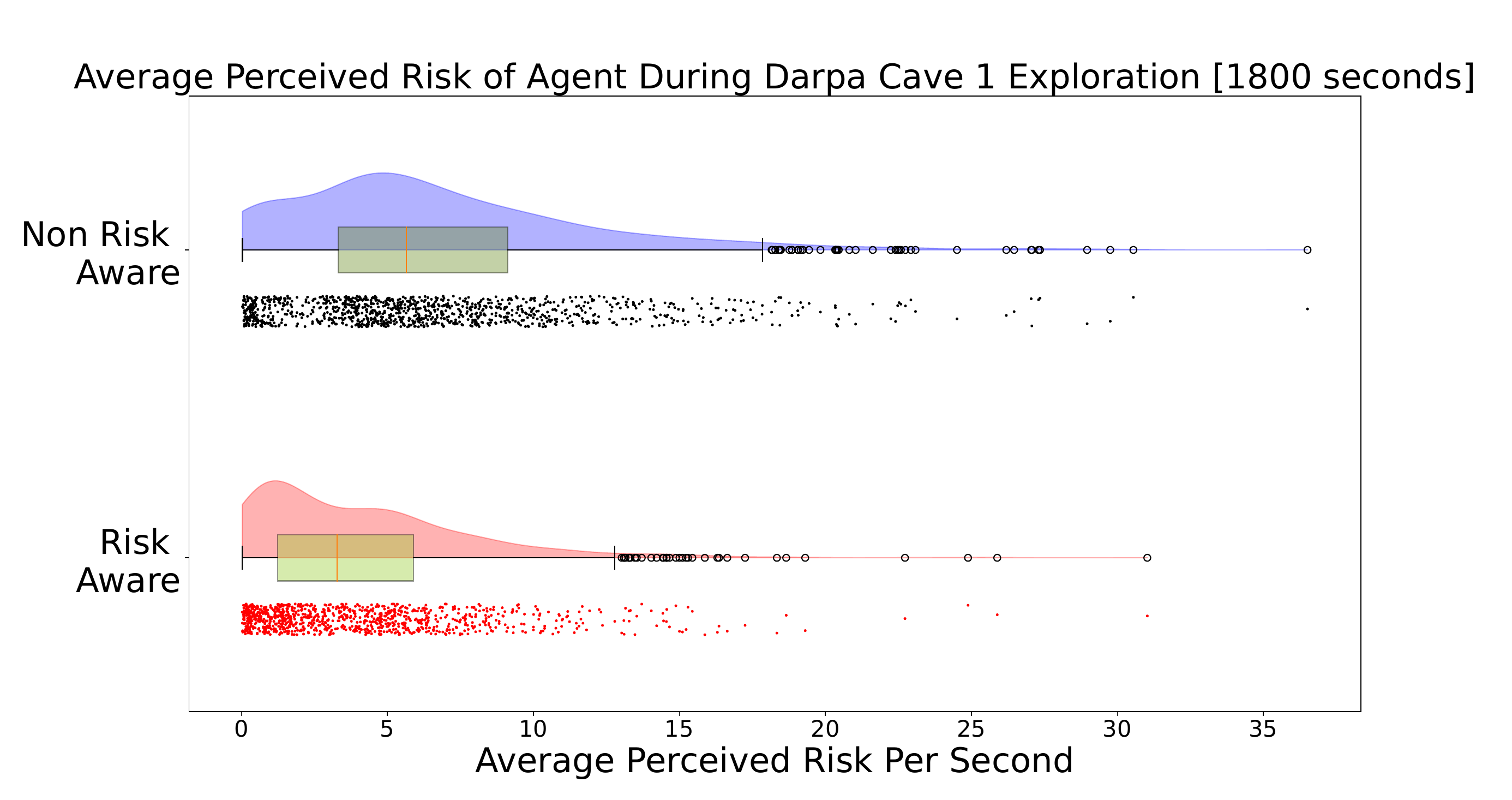}
\caption{Raincloud plots present the average Perceived Risk by the agents for 10 runs each in Darpa Cave 1 for 1800 seconds with risk-aware planning (red) and without the risk assessment framework (blue)}
    \label{fig:risk}
\end{figure}
\begin{figure}[h!]
\includegraphics[width=8.6cm]{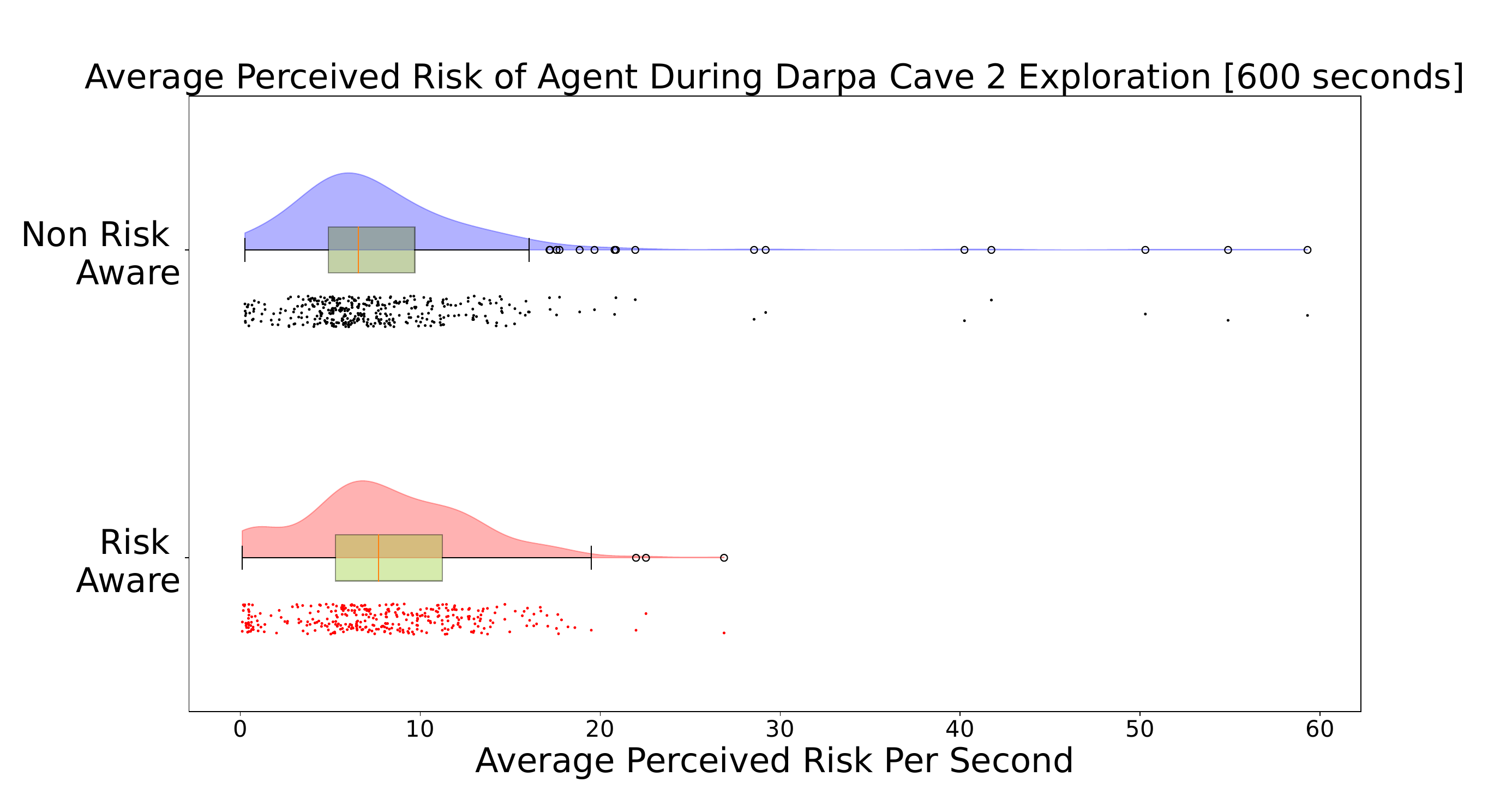}
\caption{Raincloud plots present the average Perceived Risk by the agents for 10 runs each in Darpa Cave 2 for 600 seconds with risk-aware planning (red) and without the risk assessment framework (blue)}
    \label{fig:risk2}
\end{figure}
\begin{figure}[h!]
    \centering
    \includegraphics[width=8.6cm]{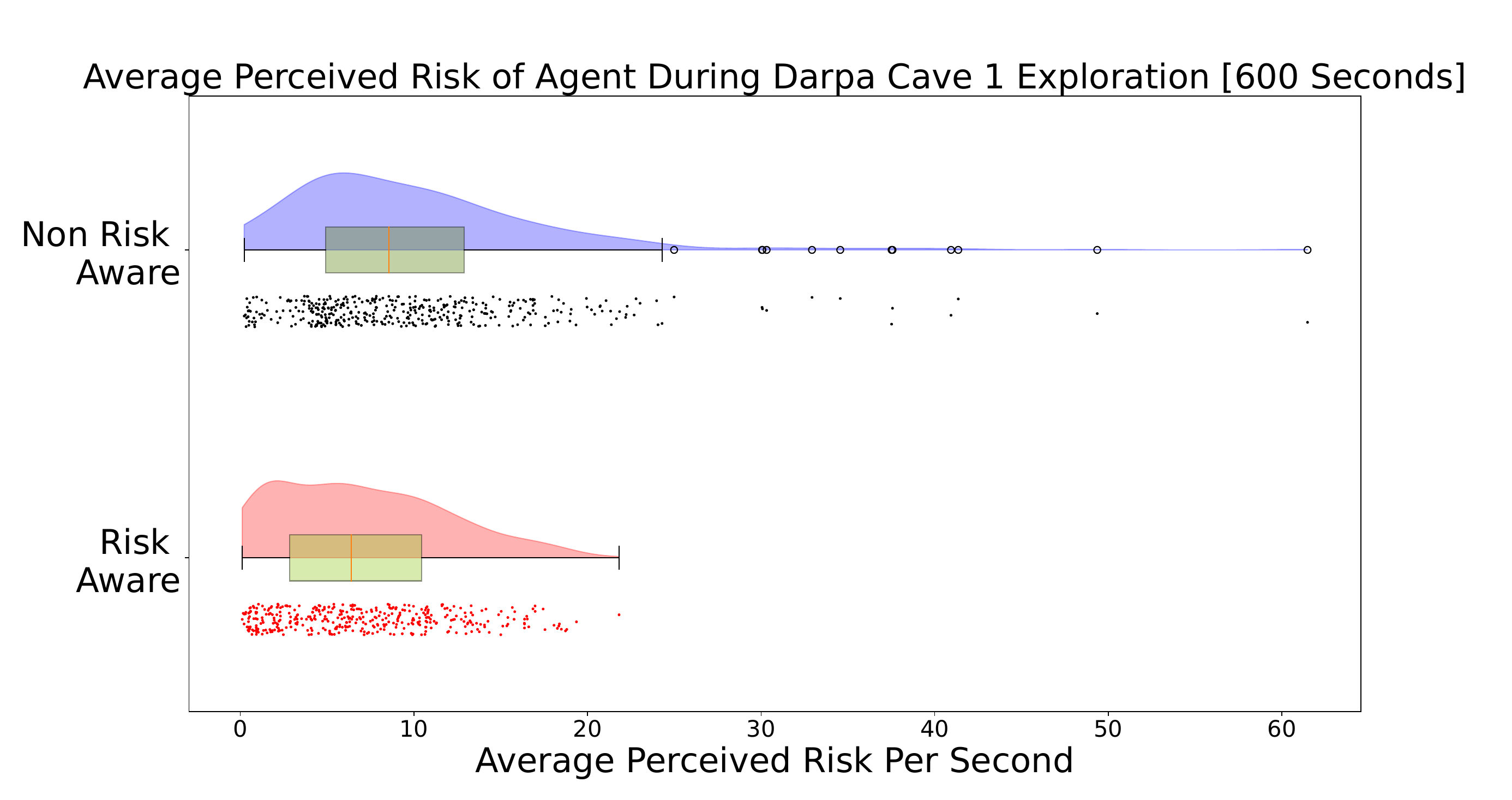}
    \caption{Raincloud plots present the average Perceived Risk by the agents for 10 runs each in Darpa Cave 1 for 600 seconds with risk-aware planning (red) and without the risk assessment framework (blue)}
    \label{fig:risk3}
\end{figure}

Figures \ref{fig:risk}, \ref{fig:risk2} and \ref{fig:risk3} present the average risk perceived by the agent per second during exploration, averaged over ten runs for each environment. The minimal difference in the distribution and mean perceived risk between runs with and without risk planning indicates that some level of risk is inherent to effectively explore highly unstructured environments. However, the outliers in the plot reveal that runs without risk planning consistently accept much higher levels of risk. This trend suggests that the agent may undertake more hazardous actions without risk management, resulting in greater variance in mission outcomes.


\begin{figure}[h!]
\includegraphics[width=8.6cm]{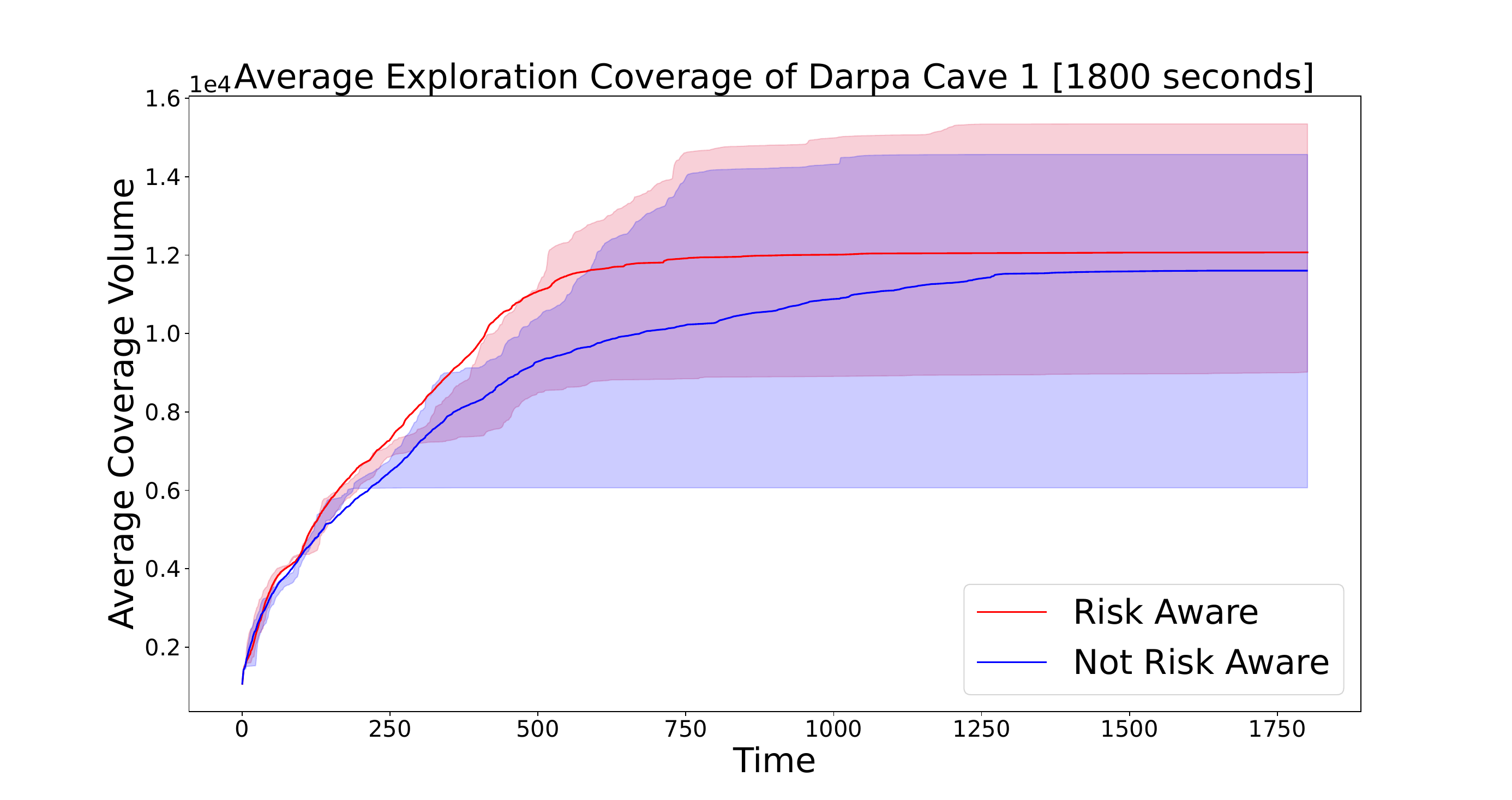}
\caption{Average Coverage Volume for ten runs Darpa Cave 1 World for 1800 seconds with and without the risk assessment framework. The area around the curve shows the distribution of data for Risk-aware planning (red) and non risk-aware planning (blue).}
    \label{fig:avgCover1}
\end{figure}
\begin{figure}[h!]
\includegraphics[width=8.6cm]{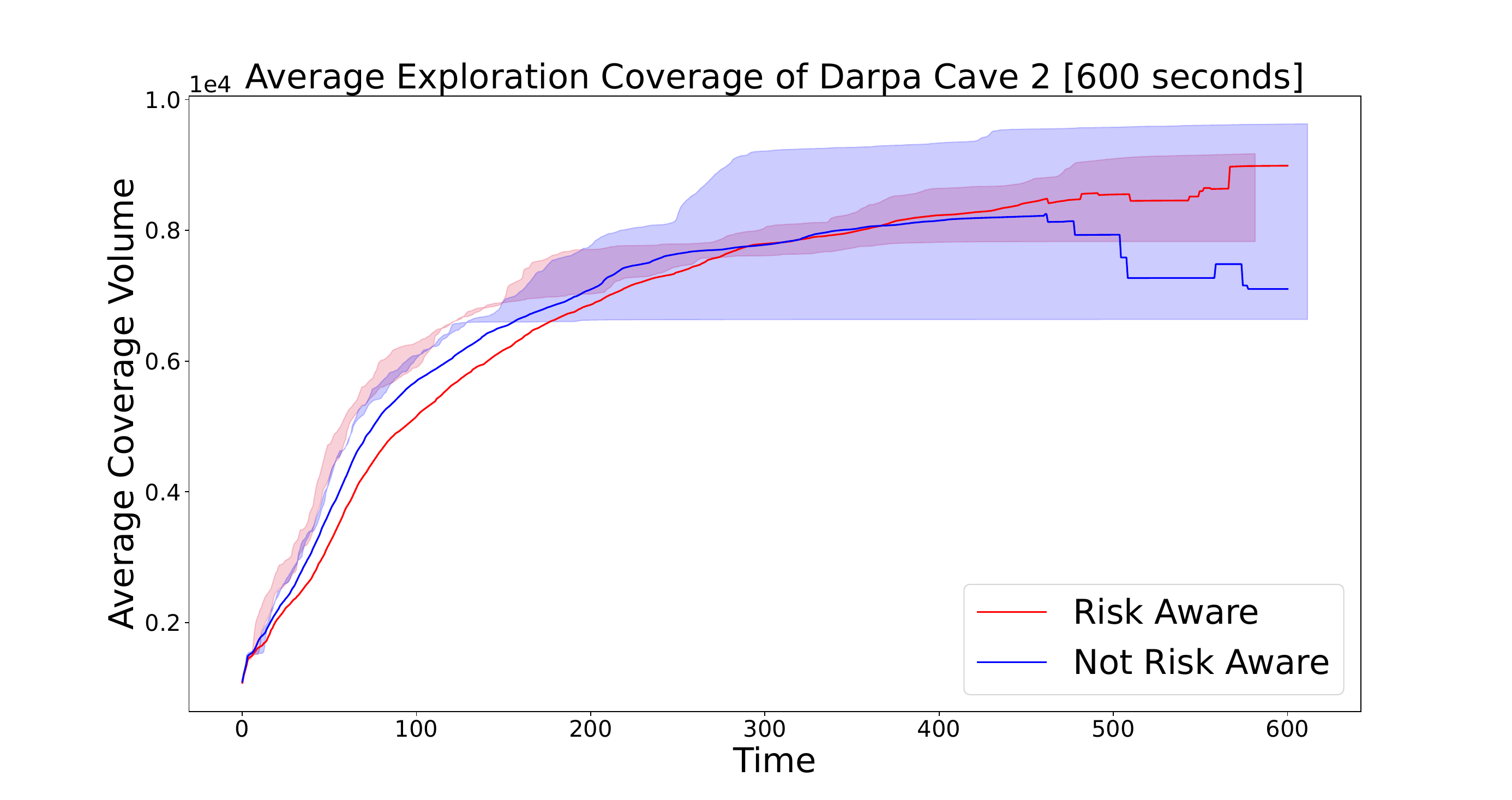}
\caption{Average Coverage Volume for ten runs for Darpa Cave 2 World for 600 seconds with and without the risk assessment framework. The area around the curve shows the distribution of data for Risk-aware planning (red) and non risk-aware planning (blue).}
    \label{fig:avgCover2}
\end{figure}

\begin{figure}[h!]
    \centering
    \includegraphics[width=8.6cm]{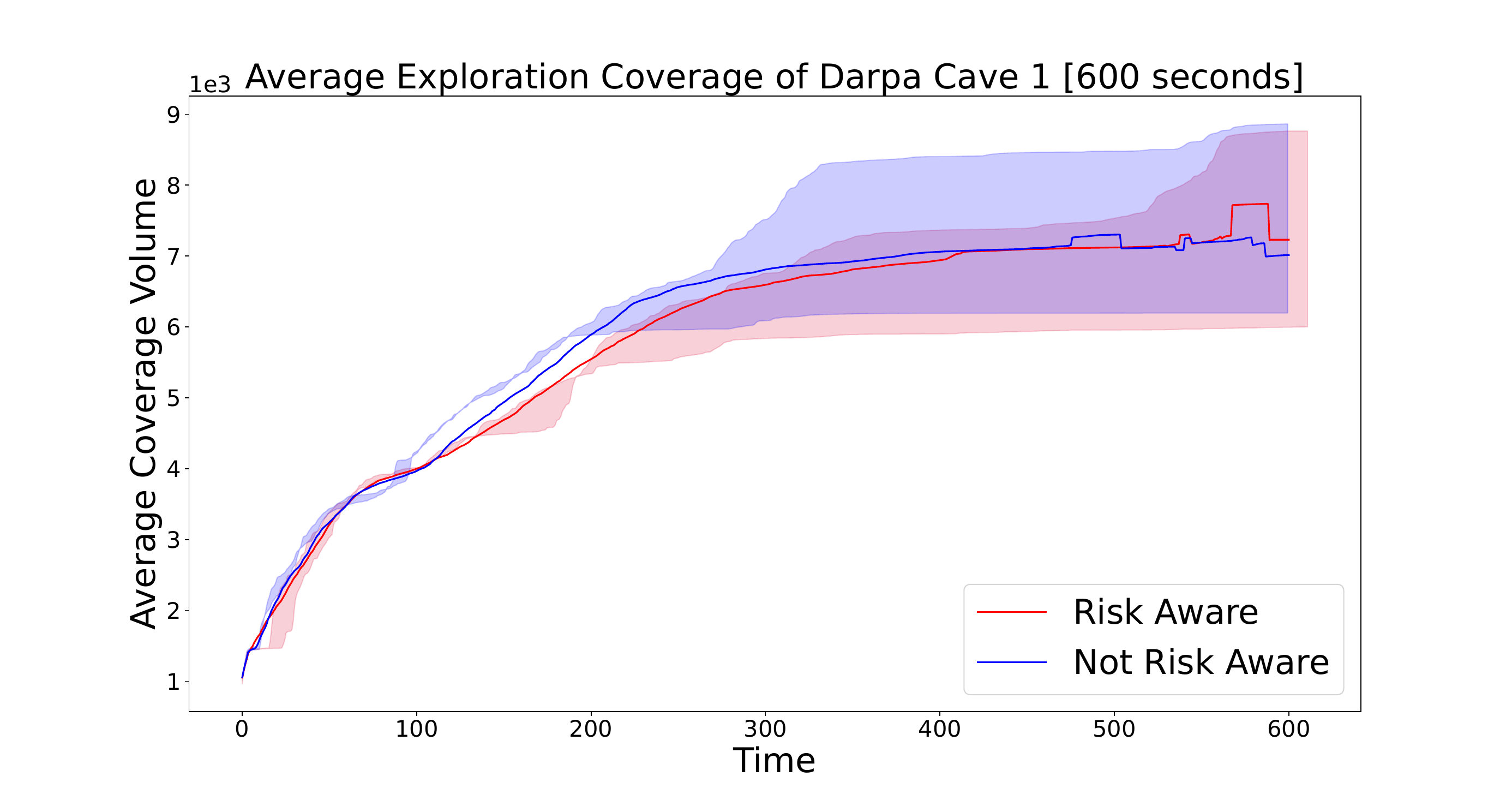}
    \caption{Average Coverage Volume for ten runs for Darpa Cave 1 World for 600 seconds with and without the risk assessment framework. The area around the curve shows the distribution of data for Risk-aware planning (red) and non risk-aware planning (blue).}
    \label{fig:avgCover3}
\end{figure}

The trend of taking more risks resulting in a larger variance in the covered area can be observed in figures \ref{fig:avgCover1}-\ref{fig:avgCover3}, and become more prominent for the experimental runs in the Darpa Cave 2 World in figure \ref{fig:avgCover2}. These figures exhibit a larger area under the curve for the explored area, indicating higher variability in the mission performance without risk planning.
The increased failure rate during non-risk-aware planning due to its inability to complete a planned action results in a drop in exploration coverage volume as the agent cannot move due to taking lethal action. Observing Figures \ref{fig:avgCover1},\ref{fig:avgCover2},\ref{fig:avgCover3}, we see a slight increase in the average coverage volume for the experiments completed with the risk assessment framework. The overall computation time to complete planning without the assessment is \textbf{0.552 seconds} while with risk planning is \textbf{0.547 seconds} indicating a negligible computational time difference. The final average increase in exploration coverage for each environment utilizing the risk assessment framework is presented in table \ref{pertable}. Incorporating the risk assessment framework into the planning process enhances the robustness of the exploration planner without increasing its computational runtime.

 \begin{table}[]
\label{pertable}
\caption{Exploration gain improvment using the risk assessment framework for each environment}
\begin{tabular}{lllll}
\cline{1-4}
\multicolumn{1}{|l|}{\textbf{Environment}}                                                         & \multicolumn{1}{l|}{\begin{tabular}[c]{@{}l@{}}Darpa Cave 1\\ {[}1800 s{]}\end{tabular}} & \multicolumn{1}{l|}{\begin{tabular}[c]{@{}l@{}}Darpa Cave 2\\ {[}600 s{]}\end{tabular}} & \multicolumn{1}{l|}{\begin{tabular}[c]{@{}l@{}}Darpa Cave 1\\ {[}600 s{]}\end{tabular}} &  \\ \cline{1-4}
\multicolumn{1}{|l|}{\textbf{\begin{tabular}[c]{@{}l@{}}Percent\\ Increase {[}\%{]}\end{tabular}}} & \multicolumn{1}{l|}{4.01}                                                                & \multicolumn{1}{l|}{8.97}                                                               & \multicolumn{1}{l|}{5.95}                                                               &  \\ \cline{1-4}
                                                                                                   &                                                                                          &                                                                                         &                                                                                         &  \\
                                                                                                   &                                                                                          &                                                                                         &                                                                                         & 
\end{tabular}
\end{table}

\section*{Conclusions and Future Work}
We propose a multi-objective risk-aware exploration planning approach that dynamically adjusts the weight of various objectives based on the current robot state and environmental conditions. The results underscore the significance of incorporating risk-aware planning during exploration. This approach adds minimal computation time while consistently performing the mission objectives.

In harsh, unstructured environments, optimizing multiple objectives often requires navigating risky paths, as efficient exploration inherently involves some level of risk. However, risk assessment enables the mitigation of lethal actions, reducing the probability of mission failure by analyzing untraversable terrain. The risk-aware approach results in reduced variance in exploration outcomes, providing greater certainty in agent performance without significantly increasing computational demands.

Future improvements could include integrating vision-based terrain analysis with the current pipeline to enhance robustness and performance. Handling sensor data remains challenging due to uncertainties associated with terrain analysis at incorrect or indeterminate points. Currently, the approach relies solely on 3D laser point cloud data. Additionally, we aim to expand this pipeline to include physical heterogeneous multi-agent systems. By considering each robot's unique capabilities during planning, the risk assessment can improve safety and decision-making in multi-agent exploration of unstructured environments.

\bibliographystyle{plain}
\bibliography{conference_101719}

\end{document}